\documentclass[sn-mathphys,Numbered]{sn-jnl}

\usepackage{graphicx}%
\usepackage{multirow}%
\usepackage{amsmath,amssymb,amsfonts}%
\usepackage{amsthm}%
\usepackage{mathrsfs}%
\usepackage[title]{appendix}%
\usepackage{xcolor}%
\usepackage{textcomp}%
\usepackage{manyfoot}%
\usepackage{booktabs}%
\usepackage{algorithm}%
\usepackage{algorithmicx}%
\usepackage{algpseudocode}%
\usepackage{listings}%
\usepackage{lscape}

\usepackage[subrefformat=parens]{subcaption}

\theoremstyle{thmstyleone}%
\theoremstyle{thmstyletwo}%

\theoremstyle{thmstylethree}%

\begin{document}

\title[Object Recognition from Scientific Document based on Compartment \& text blocks refinement Framework]{Object Recognition from Scientific Document based on Compartment \& Text Blocks Refinement Framework}


\author*[1]{\fnm{JINGHONG} \sur{LI}}\email{lijinghong-n@jaist.ac.jp}

\equalcont{These authors contributed equally to this work.}

\author*[2]{\fnm{Wen} \sur{Gu}}\email{wgu@jaist.ac.jp}

\author*[2]{\fnm{Koichi} \sur{Ota}}\email{ota@jaist.ac.jp}

\author*[2]{\fnm{Shinobu} \sur{Hasegawa}}\email{hasegawa@jaist.ac.jp}
\equalcont{These authors contributed equally to this work.}

\affil*[1]{\orgdiv{Division of Advanced Science and Technology}, \orgname{Japan Advanced Institute of Science and Technology}, \orgaddress{\street{Asahidai}, \city{Nomi}, \postcode{9231292}, \state{Ishikawa}, \country{Japan}}}

\affil*[2]{\orgdiv{Center for Innovative Distance Education and Research}, \orgname{Japan Advanced Institute of Science and Technology}, \orgaddress{\street{Asahidai}, \city{Nomi}, \postcode{9231292}, \state{Ishikawa}, \country{Japan}}}


\abstract{With the rapid development of the internet in the past decade, it has become increasingly important to extract valuable information from vast resources efficiently, which is crucial for establishing a comprehensive digital ecosystem, particularly in the context of research surveys and comprehension. The foundation of these tasks focuses on accurate extraction and deep mining of data from scientific documents, which are essential for building a robust data infrastructure. However, parsing raw data or extracting data from complex scientific documents have been ongoing challenges. Current data extraction methods for scientific documents typically use rule-based (RB) or machine learning (ML) approaches. However, using rule-based methods can incur high coding costs for articles with intricate typesetting. Conversely, relying solely on machine learning methods necessitates annotation work for complex content types within the scientific document, which can be costly. Additionally, few studies have thoroughly defined and explored the hierarchical layout within scientific documents. The lack of a comprehensive definition of the internal structure and elements of the documents indirectly impacts the accuracy of text classification and object recognition tasks. From the perspective of analyzing the standard layout and typesetting used in the specified publication, we propose a new document layout analysis framework called CTBR(Compartment \& Text Blocks Refinement). Firstly, we define scientific documents into hierarchical divisions: base domain, compartment, and text blocks. Next, we conduct an in-depth exploration and classification of the meanings of text blocks. Finally, we utilize the results of text block classification to implement object recognition within scientific documents based on rule-based compartment segmentation.}




\maketitle

\section{Introduction}\label{sec1}
Over the preceding decade, technologies opened up new fields of research and practice that center on database mining and knowledge discovery\cite{1}. With the boom in the development of Digital Ecosystems, an overwhelming number of scientific articles are available online, including Springer, ACM, IEEE Xplore, etc.\cite{2}. The open-access articles in these platforms are processed to build infrastructure databases, such as the Semantic Scholar API\cite{3} and unarXive\cite{4}, which are important for researchers and students as critical sources of advanced knowledge and reference material\cite{5}.

ChatGPT is an AI-powered conversational large language model (LLM). The potential uses of LLMs in research activities could be promising, as long as we actively consider and address any valid concerns associated with them\cite{6}. With the rise of ChatGPT, which relies on vast amounts of background data, researchers are using generative AI, including scispace\cite{7}\cite{add}, for conducting research surveys. The generated responses are mostly based on existing papers and academic data. However, the information provided to the generating AI is affected by intermittent textual information and incomplete charts, so the output of the generating AI may be incomplete, leading to low-quality generation. Therefore, To achieve higher-quality outputs from generative AI, it is crucial to accurately identify and segment content from the raw data of scientific documents.

Data mining technology is essential for constructing foundational data for research articles. As the Text analysis tools for scientific documents have been increasingly developed, which include techniques like extractive automatic summarization, abstract automatic summarization, and visualization slide generation using natural language processing\cite{8}.
In order to carry out the task mentioned above, it is important to efficiently collect and organize the language text elements (main text, itemized form, sections, footnotes) and non-language text elements (figures, tables, formulas, quotation marks) within scientific documents\cite{jsai}. For this issue, data extraction techniques are fundamental approaches to classify and identify information in the article into different categories\cite{10}. Document layout analysis (DLA) purposes to detect and annotate the physical structure of documents\cite{11}. However, parsing the layout and analyzing the content of scientific documents can be challenging and intricate. The layout of research articles is often irregular, and the typesetting styles vary depending on the publication\cite{12}. Therefore, there is a tendency to extract discontinuous data when extracting internal information from scientific documents. For instance, the flow of the main narrative from a file may be broken in mid-sentence by errors derived from the reading order of individual text blocks and interruptions, including figure titles, footnotes, and headers\cite{13}. Rule-based conditional branching and regular expressions are conventional methods to process text to solve the above issue, such as parsing sentences and identifying keywords\cite{12}. However, these methods could be improved in their ability to process the complex structure used in scientific documents, which may contain many patterns and irregularities that are not easily detected by traditional algorithms. To address this issue, researchers have turned to machine learning algorithms, which are more flexible and able to learn and adapt based on the presented data. Machine learning methods can take into account the specific patterns in scientific documents and improve the accuracy and effectiveness of automatic text processing\cite{14}. However, scientific documents may have varying fonts and typesetting, which require more complex vectorization and annotation methods to increase generality. Formulating these intricate techniques can be time-consuming and expensive\cite{TBRF}.\\
In this study, we propose the compartment \& text blocks refinement(CTBR) Framework to recognize objects from scientific documents. We utilize a rule-based method to extract single-modal blocks and rough segment the compartment based on them. Then, we use machine learning techniques to classify multi-modal text blocks that contain complex information. This helps refine the compartment and achieve object recognition in scientific documents. The contribution of the CTBR framework is mainly in the following four aspects.
\subsection{Contribution}
\begin{itemize}

    \item We propose a novel framework for understanding the layout of scientific documents in a hierarchical structure. This framework includes base domains, compartments, and text blocks, with a hierarchical structure that clearly represents the functionality of single-modal and multi-modal elements.
    \item To process text blocks, which are the fundamental elements of scientific document layout analysis in this work, we developed an integrated encoding template highlighting their characteristics. These patterns encompass dimensions, coordinates, font type, font size, and text density within the text blocks.
    \item To differentiate between the different types of information conveyed by each text block, we manually annotated the linguistic and non-linguistic information in a short period. This allowed us to create a small-scale dataset for implementing a text block classification module based on machine learning technology. Our approach is characterized by its relatively low time cost for training on specific sets of scientific documents. This enables accurate multi-modal text block classification and information extraction for large volumes of similarly formatted scientific documents.
    \item Based on the classification results, we implemented a compartment segmentation module to improve the identification of figures and tables to achieve more accurate object recognition for complex cases. In order to evaluate the effectiveness of our proposed method for object recognition, we conducted comparison experiments with existing multi-modal document processing models.
    
\end{itemize}
\section{Related work}\label{sec2}

\begin{sidewaystable*}
\caption{THE SUMMARY OF PREVIOUS WORK}
\label{t1}
\begin{tabular}{llll}
\hline
\textbf{Reference}                                                      & \textbf{Method}                                                    & \textbf{Feature/Advantage}                                                                                                                                                                                         & \textbf{Limitation}                                                                                                                                                                          \\ \hline
\textit{\begin{tabular}[c]{@{}l@{}}VGT\\ (2023)\cite{vgt}\end{tabular}}           & \begin{tabular}[c]{@{}l@{}}Grid \\ Transformer\end{tabular}        & \begin{tabular}[c]{@{}l@{}}• New diverse and detailed \\ manually-annotated dataset $D^
4LA$
 \\ •  In-depth layout analysis\\ • Pre-trained for 2D token-level \\ and segment-level semantic understanding\end{tabular} & \begin{tabular}[c]{@{}l@{}}If there is a cluster of text blocks in a \\ figure/table, it can be difficult \\ to distinguish whether it is body \\ text or figure/table region.\end{tabular} \\ \hline
\begin{tabular}[c]{@{}l@{}}Table \\ Transformer\\ (2022)\cite{table tran}\end{tabular}   & \begin{tabular}[c]{@{}l@{}}Detection \\ Transformer\end{tabular}   & \begin{tabular}[c]{@{}l@{}}Large training data, and supports multiple\\ input modalities and is useful for many\\ modeling approaches.\end{tabular}                                                                & \begin{tabular}[c]{@{}l@{}}Miss detection in complex pattern \\ matching involving continuous tables.\end{tabular}                                                                           \\ \hline
\textit{\begin{tabular}[c]{@{}l@{}}Table net\\ (2020)\cite{table net}\end{tabular}}     & \begin{tabular}[c]{@{}l@{}}End-to-end\\ Deep learning\end{tabular} & \begin{tabular}[c]{@{}l@{}}This model uses interdependence \\ between 0table detection and \\ structure recognition tasks to \\ segment table and column regions.\end{tabular}                                     & Header of the table is difficult to fit.                                                                                                                                                     \\ \hline
\textit{\begin{tabular}[c]{@{}l@{}}Pdffigure 2.0\\ (2016)\cite{pdffigure}\end{tabular}} & Rule-based                                                         & \begin{tabular}[c]{@{}l@{}}Analyzes page structure and locates \\ figures and tables by analyzing empty \\ regions within text.\end{tabular}                                                                       & \begin{tabular}[c]{@{}l@{}}Miss detection occurred when figures\\ and tables appear continuously.\end{tabular}                                                                               \\ \hline
\textit{\begin{tabular}[c]{@{}l@{}}Tabula\\ (2018)\cite{tabula}\end{tabular}}        & Rule-based                                                         & \begin{tabular}[c]{@{}l@{}}It uses customizable heuristics to \\ detect tables and reconstruct \\ cell structure based on text and \\ ruling lines in the \textit{PDF}.\end{tabular}                                        & \begin{tabular}[c]{@{}l@{}}If recognition is not restricted to the \\ table zone, the body-text and \\ section title patterns may be mixed \\ with the detected result.\end{tabular}         \\ \hline
\textit{\begin{tabular}[c]{@{}l@{}}Grobid\\ (2008-2023)\cite{grobid}\end{tabular}}   & CRF                                                                & \begin{tabular}[c]{@{}l@{}}Extract and reorganize not only \\ the content but also the layout \\ and text styling information.\end{tabular}                                                                        & \begin{tabular}[c]{@{}l@{}}It is difficult to extract noise-free text\\ because information with figures, tables,\\ and equations are included in the body-text.\end{tabular}                \\ \hline
\end{tabular}
\end{sidewaystable*}

\subsection{Existing research on layout analysis of scientific document \textit{PDFs}}
Cheng et al. present a two-stream multi-modal Vision Grid Transformer for document layout analysis, which directly models 2D token-level and segment-level semantic understanding\cite{vgt}. 
Lopez et al. developed GROBID, which extracts the bibliographical data corresponding to the header information (title, authors, abstract, etc.) and to each reference (title, authors, journal title, issue, number, etc.)\cite{grobid}.

\subsection{Recognition of figures and tables in scientific article \textit{PDFs}}
Clark et al. developed PDFFigures 2.0, which extracts figures, tables, and captions from computer science articles in \textit{PDF} format. The algorithm analyzes the structure of individual pages by detecting captions, graphical elements, and chunks of body-text and then locates figures and tables by reasoning about the empty regions within that text\cite{pdffigure}.
Frerebeau et al. provided Tabula, a web-based system that extracts tables from untagged \textit{PDF} documents. It uses customizable heuristics to detect tables and reconstruct cell structure based on text and ruling lines in the \textit{PDF}\cite{tabula}.


\subsection{Recognition of figures and tables in images of various formats}

Smock et al. created PubTables-1M, a new dataset for table extraction from scientific articles. It has almost one million tables, detailed headers, and location information for table structures. PubTables-1M also supports multiple input modalities and is useful for many modeling approaches. Their transformer-based object detection models trained on PubTables-1M produce excellent results for detection, structure recognition, and functional analysis without any special customization\cite{table tran}.
Paliwal et.al built a novel end-to-end deep learning model for both table detection and structure recognition. The model exploits the interdependence between the table detection and table structure recognition to segment the table and column regions\cite{table net}.

\subsection{Position of this work}
The \textbf{Table \ref{t1}} presents a summary of the previous research and their limitations on data extraction for scientific documents. Previous studies have faced difficulty in accurately distinguishing text blocks within figures and tables. This study aims to address these limitations by refining text blocks using different features, including position, size, line and column spacing, font type, and font size. By considering these features and compartment segmentation, we can enhance the ability to distinguish various objects within scientific documents.

\section{Methodology-Definition}
\subsection{Overview}
\begin{figure}[htbp]
\centering
\includegraphics[width=14cm,height=8cm]{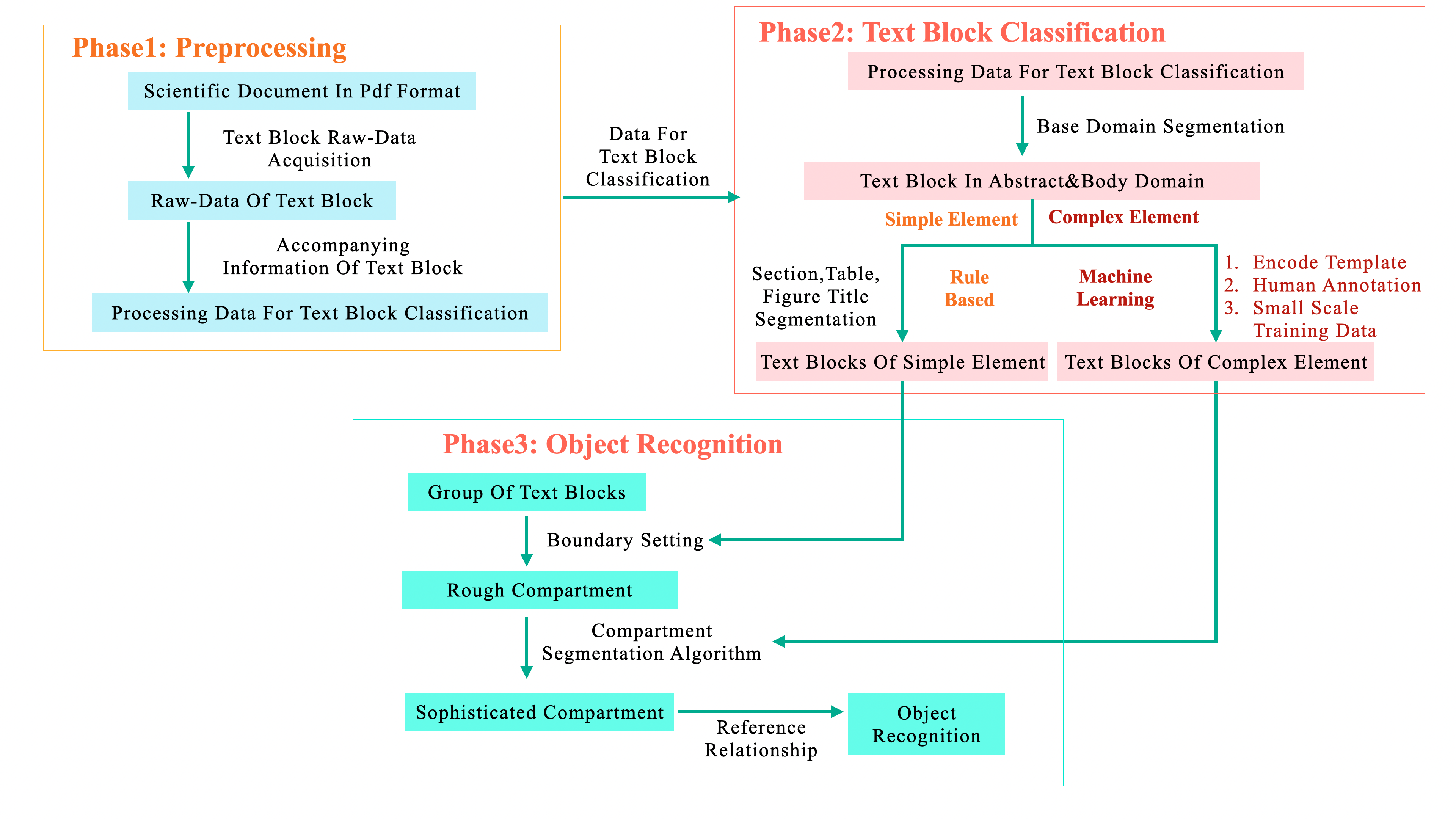}
\caption{Overview}
\label{f1}
\end{figure}

We are dedicated to creating a framework for document layout analysis using text block and compartment analysis. Our approach utilizes rule-based algorithms to process text within identifiable text blocks to implement rough compartment segmentation. Additionally, we employ machine learning models for multi-modal text block classification. Furthermore, by combining the classification result with rough compartment segmentation, we use a sophisticated compartment refinement algorithm to achieve object recognition. Our framework consists of three stages, as illustrated in \textbf{Fig.\ref{f1}}.\\

\begin{itemize}
\item The first stage is preprocessing, which provides detailed instructions on parsing \textit{PDFs} and extracting information from text blocks from scientific documents.\\
\item The second stage involves text block classification, using a combination of rule-based methods to identify single-modal text blocks and machine learning to classify complex text blocks.\\
\item The third stage focuses on the algorithm of compartment segmentation and object recognition for figures and tables based on the results of text block classification in the second stage. \\
\end{itemize}

\subsection{Internal Environment of scientific document}
\begin{figure}[htbp]
\centering
\includegraphics[width=13cm,height=8cm]{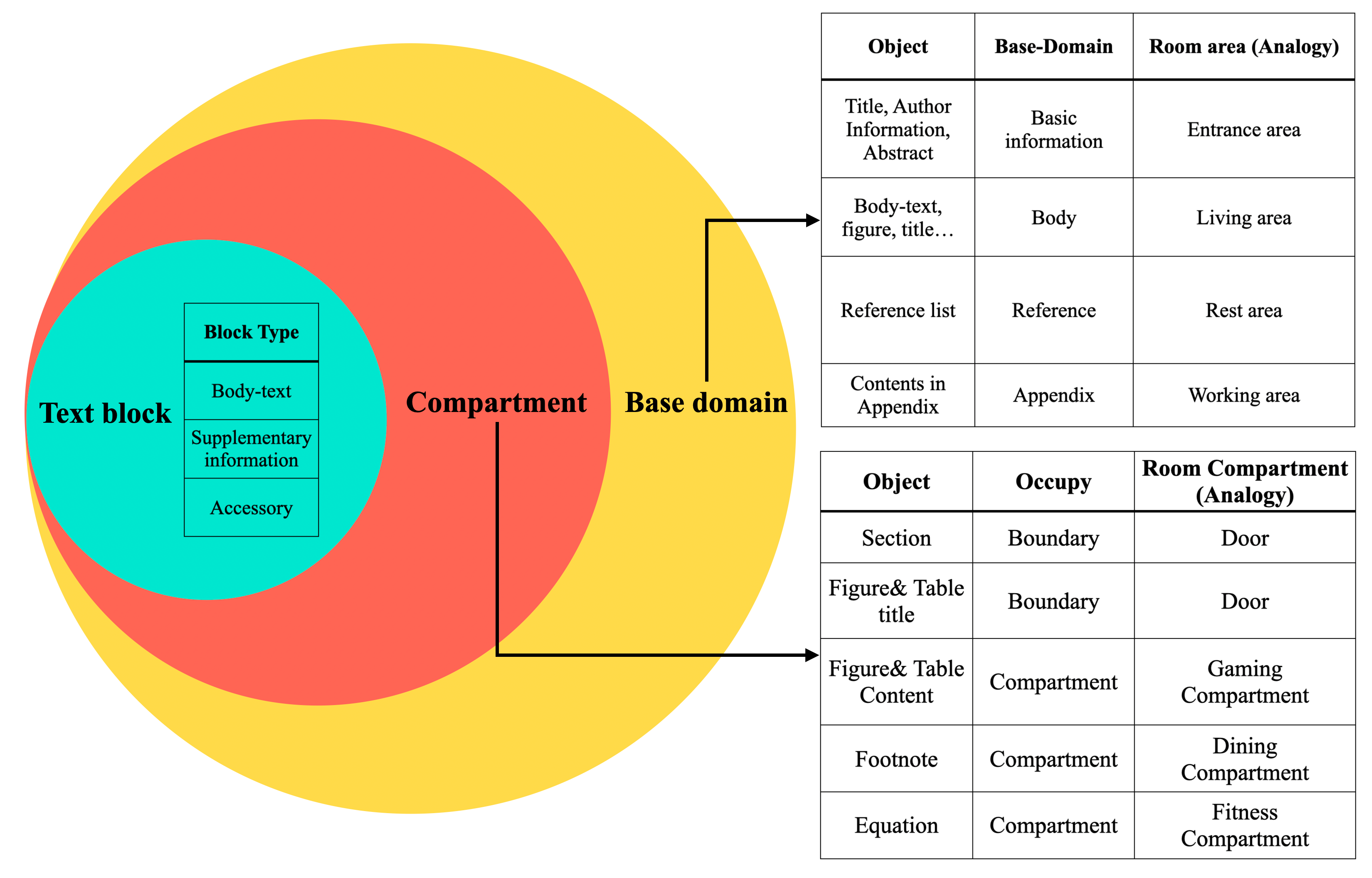}
\caption{Internal Environment of scientific document}
\label{f2}
\end{figure}

To gain a deeper understanding of document layout, it is important to define the Internal Environment in the document for developing a simulation model, similar to urban planning or apartment room layout design. In this study, we categorize the internal structure of scientific documents into three levels. In this section, we provide specific definitions for the base domains, compartments, and text blocks within the context of scientific documents, as well as the roles they play in our overall system. The intuitive hierarchical structure is shown in \textbf{Fig.\ref{f2}}.\\

\begin{itemize}
\item 1. The first level consists of base domains that form the overall structure, such as basic information domain (the region before the chapter 1 included title, author information, mail address, abstract.etc.),  Its function is to provide readers about the meta information of this article and a general understanding. Comparable to the entrance area inside a house, when you step into the entrance area of a house, you may see the overall style and structure of the house, such as the color scheme, the general layout of the rooms, and the decorative style. This will give you a rough impression of the house.\\
\item 2. The second level further subdivides each base domain into compartments, representing the different functions they serve. For instance, a model diagram can provide readers with an overview of research methods, helping them organize their thoughts and injecting energy into their understanding of the documents. Comparable to the dining compartment , which provide people with the necessary energy for their daily lives that promote engagement in more activities within the house.\\
\item 3. The third level comprises text block within the compartments, which are the components that make up the compartments. Each text block also has specific functions that contribute to the overall function of the compartment. For instance, in a table, some text blocks contain data of accuracy rate that gives readers an intuitive expression to help them understand the intended purpose of the table as conveyed by the author. Comparable to the furniture in the dining compartment, such as dining tables, chairs, and wine cabinets, these furniture items are physical objects that we use in our dining process. They provide us a more tangible experience of the dining scene.\\
\end{itemize}

\subsubsection{Text block in scientific document}
The text block is a collection of different character strings. Just like authors organize scientific documents, they often group strings that express specific contents for formatting purposes. When these strings are closely arranged, they tend to form groups. Some text blocks feature single-modal expressions that convey singular and easily processed information, such as figure titles, table titles and section titles. These can be classified using rule-based and regular expression methods. However, rule-based methods do not easily understand some text blocks, particularly those embedded in figures/tables and separated from the body-text. These multi-modal text blocks appearing in tables may represent measurements of experimental evaluation criteria, while those appearing in figures may describe components of research modules. These text blocks serve different purposes than similar-looking text blocks in the body-text. In this study, we aim to use machine learning methods for feature recognition of multi-modal text blocks. The types of text blocks are summarized in \textbf{Table \ref{t2}}. To enable machines to analyze text blocks containing multi-modal information, we categorized the usage of text blocks in scientific documents that carry multi-modal information into three categories: Body-text, Supplementary information, and Accessory information\cite{TBRF}. Their definitions are shown in \textbf{Table \ref{t3}}.

\begin{table}[htbp]
\caption{Type \& characteristics of Object}
\label{t2}
\begin{tabular}{|l|c|l|c|}
\hline
\textbf{Object}                                                           & \multicolumn{1}{l|}{\textbf{Type}} & \textbf{Reason}                                                                                                                          & \multicolumn{1}{l|}{\textbf{Method}} \\ \hline
\textit{\textbf{\begin{tabular}[c]{@{}l@{}}Section \\ Title\end{tabular}}} & Obvious                           & \begin{tabular}[c]{@{}l@{}}• Continuity of section numbers \\ • Specific fonts for section titles\end{tabular}                           & RB                                   \\ \hline
\textit{\textbf{\begin{tabular}[c]{@{}l@{}}Tab\&Fig\\ Title\end{tabular}}} & Obvious                           & \begin{tabular}[c]{@{}l@{}}The format of Tab/Fig titles is \\ generally consistent within the \\ same academic publication.\end{tabular} & RB                                   \\ \hline
\textit{\textbf{Body-Text}}                                                & Unobvious                          & Discontinuous data                                                                                                                       & ML                                   \\ \hline
\textit{\textbf{Figure}}                                                   & Unobvious                          & Irregular text block included                                                                                                            & ML                                   \\ \hline
\textit{\textbf{Table}}                                                    & Unobvious                          & Irregular text block included                                                                                                            & ML                                   \\ \hline
\textit{\textbf{Page\_Num}}                                                & Unobvious                          & Similar text block in tables                                                                                                             & ML                                   \\ \hline
\textit{\textbf{Footnote}}                                                 & Unobvious                          & Similar text block in body-text                                                                                                          & ML                                   \\ \hline
\end{tabular}
\end{table}

\begin{table}[htbp]
\centering
\caption{Type of text block}
\label{t3}
\begin{tabular}{|c|c|}
\hline
\textbf{Type}                                                       & \textbf{Disciption}                                                                                                                     \\ \hline
\textbf{Body-text}                                                            & Sentence group in body of article                                                                                                    \\ \hline
\textbf{\begin{tabular}[c]{@{}c@{}}Supplementary\\  information\end{tabular}} & \begin{tabular}[c]{@{}c@{}}Figure and Table regions, \\ Figure and Table titles  \\ Equations, Algorithms, Sections title\end{tabular} \\ \hline
\textbf{Accessory information}                                                & \begin{tabular}[c]{@{}c@{}}Page number, Footnote, \\ Meta information\end{tabular}                                                   \\ \hline
\end{tabular}
\end{table}

\subsection{Definition of Compartment}
This study defines a compartment in scientific documents as a group of multiple text blocks. These compartments provide a richer information combination than a single text block. For instance, when we see a figure in an article, we need to integrate the relationships between different parts of the figure to understand them fully. In this case, the figure region can be viewed as a compartment, where the information of different parts is often presented through text blocks or graphic elements. Therefore, accurately segmenting and recognizing the content inside compartments and the information they convey is crucial for improving machine perception of the document layout. In section 4.4, we explain in detail how to utilize the results of text block classification for object detection based on a compartment segmentation algorithm.

\section{Methodology-implementation}
\subsection{Phase1:Preprocessing}
\subsubsection{Text block parsing}
\begin{figure}[htbp]
\centering
\includegraphics[width=14cm,height=9.5cm]{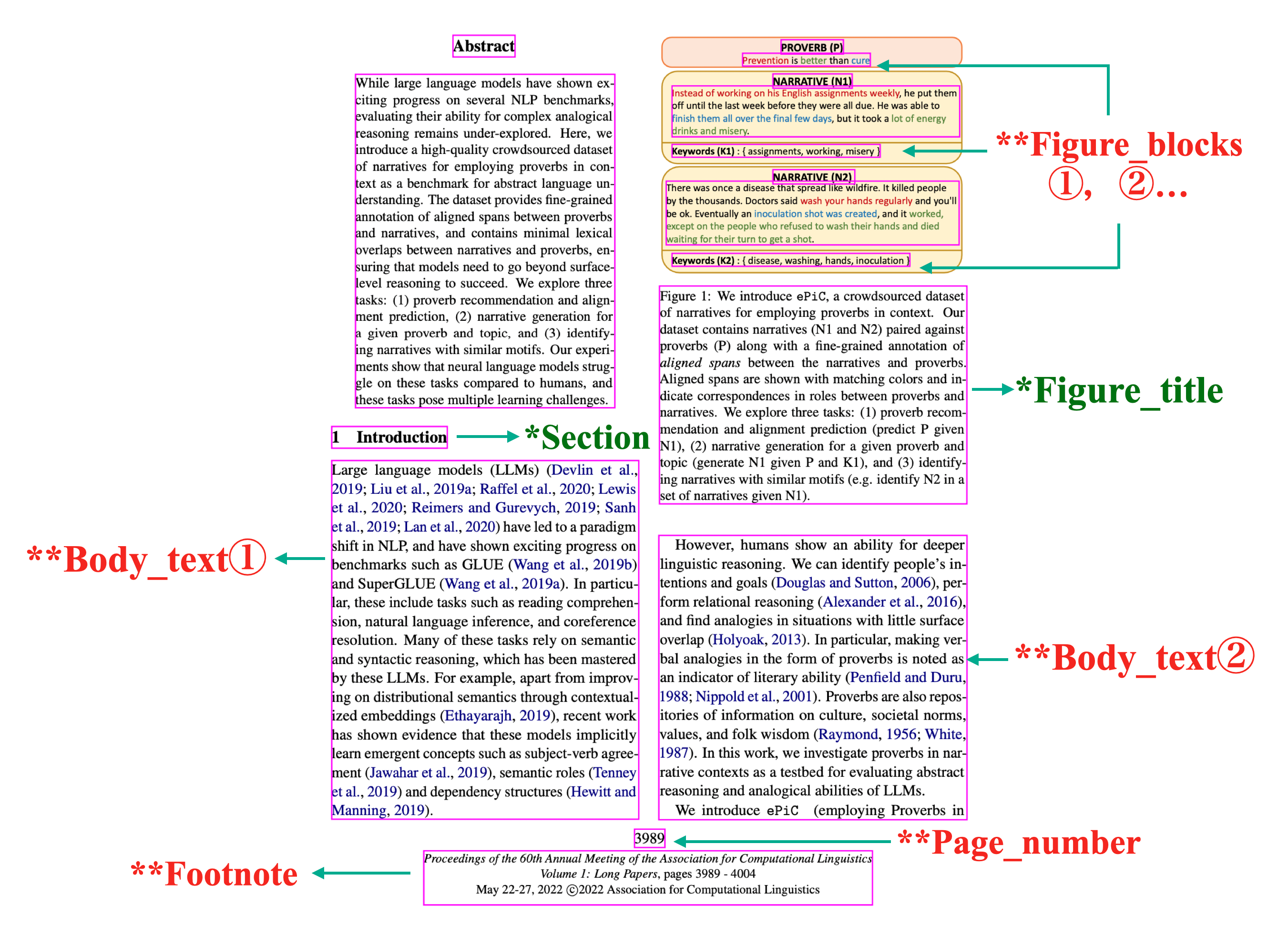}
\caption{Unstructured page layout: Sample article\cite{fig3}}
\label{f3}
\end{figure}

Scientific repositories currently store research articles in \textit{PDF} format. The first step is to focus on parsing the text blocks within \textit{PDF} articles. This can be automated using the external library \textit{pymupdf}\cite{pymupdf} in Python. By analyzing the raw structure of scientific documents, the line and column spacing, characteristics can be used to divide the text of a \textit{PDF} file into multiple sub-text blocks.These blocks contain specific content, and their combination results compose the unstructured page layout, as depicted in \textbf{Fig.\ref{f3}}. The Page.get text(“blocks”) method of \textit{pymupdf} can be used to extract the text blocks of each page.

\subsubsection{Extracting accompanying information from text}
When extracting text blocks using \textit{pymupdf}, the accompanying text’s information, such as font, size, and style, can also be obtained\cite{pymupdf}. This information can be used in various ways, such as analyzing the document’s structure, identifying headers and footers, and detecting text requiring special formatting. Analyzing these characteristics makes it possible to optimize the extraction process and improve the accuracy of extracting specific features. For example, font size can be used to identify the main-sections of the document or to differentiate between headings and body-text. Font style can be used to detect emphasis or to identify quotes. Moreover, utilizing the accompanying text’s information enables data extraction in a more organized and structured manner, resulting in more easily obtainable, low-noise output. The accompanying information extracted in this study is shown in \textbf{Table \ref{t4}}. 

\begin{table}[]
\caption{Accompanying information and their usage}
\label{t4}
\begin{tabular}{cl}
\hline
\textbf{Accompanying information} & \multicolumn{1}{c}{\textbf{Usage}}                                                                       \\ \hline
Bounding box                      & Encode for left, right, top, bottom, width,height                                                        \\ \hline
Font size                         & \begin{tabular}[c]{@{}l@{}} • Boundary setting\\  • Encode for font size, text blockdensity\end{tabular} \\ \hline
Font type                         & \begin{tabular}[c]{@{}l@{}} • Boundary setting\\  • Encode for font type\end{tabular}                    \\ \hline
Media box                         & Compartment segmentation                                                                                 \\ \hline
\end{tabular}
\end{table}

\subsection{Phase 2-1: Rule-based Implementation for simple text block element}
\subsubsection{Base domain segmentation}
We defined the Basic Information Domain, Body Domain, Reference Domain, and Appendix Domain as the fundamental domains. The subsequent work of this research specifically focuses on the Body Domain. We utilize regular expressions and text accompanying information within the text blocks to perform compartment segmentation. 

\subsubsection{Single-modal text block recognition}
This study focuses on processing scientific documents that have specific section numbers assigned. There are two types of section titles: main-section and sub-section. \textbf{Table \ref{t6}} shows each matching method using regular expression. The font characteristics(font type and font size) of the text are also utilized to distinguish similar patterns in the body-text. Figure and table titles use a similar recognition method, as shown in \textbf{Table \ref{t6}}.

\begin{table}[]
\caption{Regular expression for single-modal text block recognition in \textit{ACL} format}
\label{t6}
\begin{tabular}{|l|l|}
\hline
\textbf{Element}  & \textbf{Regular expression}                                                                                                              \\ \hline
Main section title     & \verb/^[0-9]{1,2}\s*[\.|,].*$/                                                 \\ \hline
Sub section title & \verb/^[0-9]{1,2}(\.[0-9]{1,2}){1,4}\s+.*$/                               \\ \hline
Figure title      & \verb/^[F|f][I|i][G|g][U|u][R|r][E|e]\s*\d+\s*:.*$/ \\ \hline
Table title       & \verb/^[T|t][A|a][B|b][L|l][E|e]\d+\s*:.*$/          \\ \hline
\end{tabular}
\end{table}

\subsection{Phase 2-2: Classification for complex text block element}
In contrast to the processing method described in the previous section for single-modal text blocks, effectively classifying the information contained in multi-modal text blocks is challenging using rule-based algorithms. This is because the conditions required for rule-based processing are complex, making it difficult to capture individual cases. This chapter extracts features from complex element's text blocks and encodes them to address the issue of classifying multi-modal information. The SVM method in machine learning is then used to recognize these feature patterns for comprehensive classification.

\subsubsection{Encoder template}
Nine vector elements are selected to create a vector that reflects the characteristics of an article. Each vector element is constructed to embed the characteristics of a specific object. The encoding method for each element is as follows:\\

\textbf{1-2, Left\_Position and Right\_Position: }
As the body-text block is often Justify Align, the goal is to accurately calculate the location of the beginning of each block and obtain common characteristics of the text. Expressly, we set the Left/Right-aligned block as a left/right boundary line and calculate the division between each block's left/Right coordinate and the boundary coordinate.\\
\begin{equation}
\small
    Code_{(left|right)} = \frac{Block\_coordinate_{(Left|Right)}}{Boundary\_coordinate_{(Left|Right)}} 
\end{equation}

\textbf{3-4, Top\_Position and Bottom\_Position: }
Footnote and page number information is often displayed at the bottom of the page, so it is necessary to encode that information to distinguish it from other information. The encoding method is the same as the one described in Left /Right Position but switches from left/right to upper/lower.\\
\begin{equation}
\small
    Code_{(top|bottom)} = \frac{Block\_coordinate_{(top|bottom)}}{Boundary\_coordinate_{(top|bottom)}} 
\end{equation}

\textbf{5-6, Width\_Length and Height\_Length: }
While the blocks of body-text are distributed more regularly, it is necessary to grasp the width and height characteristics of the blocks to recognize the supplement information blocks, as they are irregularly distributed and have many small blocks. Therefore, in an article, set the largest width/height as the standard for all blocks, perform a division of the width/height of each block, and encode it.\\
\begin{equation}
\small
    Code_{(width|height)} = \frac{Block\_size_{(width|height)}}{Max\_size_{(width|height)}} 
\end{equation}

\textbf{7, Font\_Type(\textbf{ft}): }
In most cases, the body-text font has the highest frequency in scientific documents. Therefore, following the font acquisition method in section \textbf{4.1.2}, all blocks are scanned to aggregate the corresponding character count for each font, and the font type with the highest frequency is considered the body-text font. Next, the most frequently appearing font in each block is calculated, and if it is the body-text font, it is encoded as' 1'. Otherwise, it is encoded as' 0'.\\

\let\saveeqnno\theequation
\let\savefrac\frac
\def\dispfrac{\displaystyle\savefrac}
\begin{eqnarray}
\let\frac\dispfrac
\gdef\theequation{4.1}
\let\theHequation\theequation
\label{dfg-2e952039a203}
\begin{array}{@{}l}\begin{array}{l}Body_{font}=Index\_of\{Max\{\sum_{}^{}ft1,...,\sum_{}^{}ftN\}\}\\\end{array}\end{array}
\end{eqnarray}
\global\let\theequation\saveeqnno
\addtocounter{equation}{-1}\ignorespaces 
\vskip-1.5\baselineskip 
\let\saveeqnno\theequation
\let\savefrac\frac
\def\dispfrac{\displaystyle\savefrac}
\begin{eqnarray}
\let\frac\dispfrac
\gdef\theequation{4.2}
\let\theHequation\theequation
\label{dfg-e0100a3ed3d4}
\begin{array}{@{}l}Code_{ft}=\left\{\begin{array}{l}\mathbf1\boldsymbol\;(Body\_font)\\\mathbf0\boldsymbol\;(Others)\end{array}\right.\end{array}
\end{eqnarray}
\global\let\theequation\saveeqnno
\addtocounter{equation}{-1}\ignorespaces 

\textbf{8, Font\_Size(\textbf{fs}): }
First, set the standard font size to the most frequently appeared font size among the body fonts. Next, determine the most frequent font size in each block and set it as the font size for that block. Finally, divide each block's font size by the standard font size and encode it. In equation (5.1 - 5.3), \textbf{fss} means font size in that text block.\\

\let\saveeqnno\theequation
\let\savefrac\frac
\def\dispfrac{\displaystyle\savefrac}
\begin{eqnarray}
\let\frac\dispfrac
\gdef\theequation{5.1}
\let\theHequation\theequation
\label{dfg-72e38e6b6e6a}
\begin{array}{@{}l}\begin{array}{l}Body_{fs}=Index\_of\{Max\{\sum_{}^{}f_s1,...,\sum_{}^{}f_sN\}\}\\\end{array}\end{array}
\end{eqnarray}
\global\let\theequation\saveeqnno
\addtocounter{equation}{-1}\ignorespaces 
\vskip-1.5\baselineskip 
\let\saveeqnno\theequation
\let\savefrac\frac
\def\dispfrac{\displaystyle\savefrac}
\begin{eqnarray}
\let\frac\dispfrac
\gdef\theequation{5.2}
\let\theHequation\theequation
\label{dfg-6133c8488da1}
\begin{array}{@{}l}Block_{fs}=Index\_of\{Max\{\sum_{}^{}f_{ss}1,...,\sum_{}^{}f_{ss}N\}\}\end{array}
\end{eqnarray}
\global\let\theequation\saveeqnno
\addtocounter{equation}{-1}\ignorespaces 
\vskip-1.5\baselineskip 
\let\saveeqnno\theequation
\let\savefrac\frac
\def\dispfrac{\displaystyle\savefrac}
\begin{eqnarray}
\let\frac\dispfrac
\gdef\theequation{5.3}
\let\theHequation\theequation
\label{dfg-2e3da6a47c31}
\begin{array}{@{}l}\begin{array}{l}Code_{fs}=\frac{Block_{fs}}{Body_{fs}}\\\end{array}\end{array}
\end{eqnarray}
\global\let\theequation\saveeqnno
\addtocounter{equation}{-1}\ignorespaces 

\textbf{9, Density of text block(\textbf{D}): }
In contrast to the basic characteristics of a text block mentioned earlier, we propose the concept of a higher-dimensional feature of text block density. This feature is defined as the ratio of the occupied area of text within a text block to the size of the text block. A larger ratio indicates a smaller blank area within the text block. Typically, body-text containing intensive text arrangement has a small occupied area of blank space. On the other hand, text blocks corresponding to table information tend to have a larger blank area due to the presence of spaces and empty lines. Therefore, we calculate the text block density using equation (6.1 - 6.3), which combines several characteristics mentioned in the previous section to determine the category of a text block.

\let\saveeqnno\theequation
\let\savefrac\frac
\def\dispfrac{\displaystyle\savefrac}
\begin{eqnarray}
\let\frac\dispfrac
\gdef\theequation{6.1}
\let\theHequation\theequation
\label{dfg-d8008613ac06}
\begin{array}{@{}l}Code_{density}=\frac{Area\;of\;text\_block}{Area\;of\;text\;occupy}=\frac{S_{block}}{Length_{text}\ast Code_{fs}}\end{array}
\end{eqnarray}
\global\let\theequation\saveeqnno
\addtocounter{equation}{-1}\ignorespaces 
\vskip-1.5\baselineskip 
\let\saveeqnno\theequation
\let\savefrac\frac
\def\dispfrac{\displaystyle\savefrac}
\begin{eqnarray}
\let\frac\dispfrac
\gdef\theequation{6.2}
\let\theHequation\theequation
\label{dfg-72c24fc630fa}
\begin{array}{@{}l}S_{block}=Width\;of\;Block_{size}\ast Height\;of\;Block_{size}\end{array}
\end{eqnarray}
\global\let\theequation\saveeqnno
\addtocounter{equation}{-1}\ignorespaces 
\vskip-1.5\baselineskip 
\let\saveeqnno\theequation
\let\savefrac\frac
\def\dispfrac{\displaystyle\savefrac}
\begin{eqnarray}
\let\frac\dispfrac
\gdef\theequation{6.3}
\let\theHequation\theequation
\label{dfg-132851d036b6}
\begin{array}{@{}l}Length_{text}=Count\;of\;text\;in\;block\;\end{array}
\end{eqnarray}
\global\let\theequation\saveeqnno
\addtocounter{equation}{-1}\ignorespaces 

\subsubsection{Annotation for text block}
\begin{figure}[htbp]
\centering
\includegraphics[width=14cm,height=11cm]{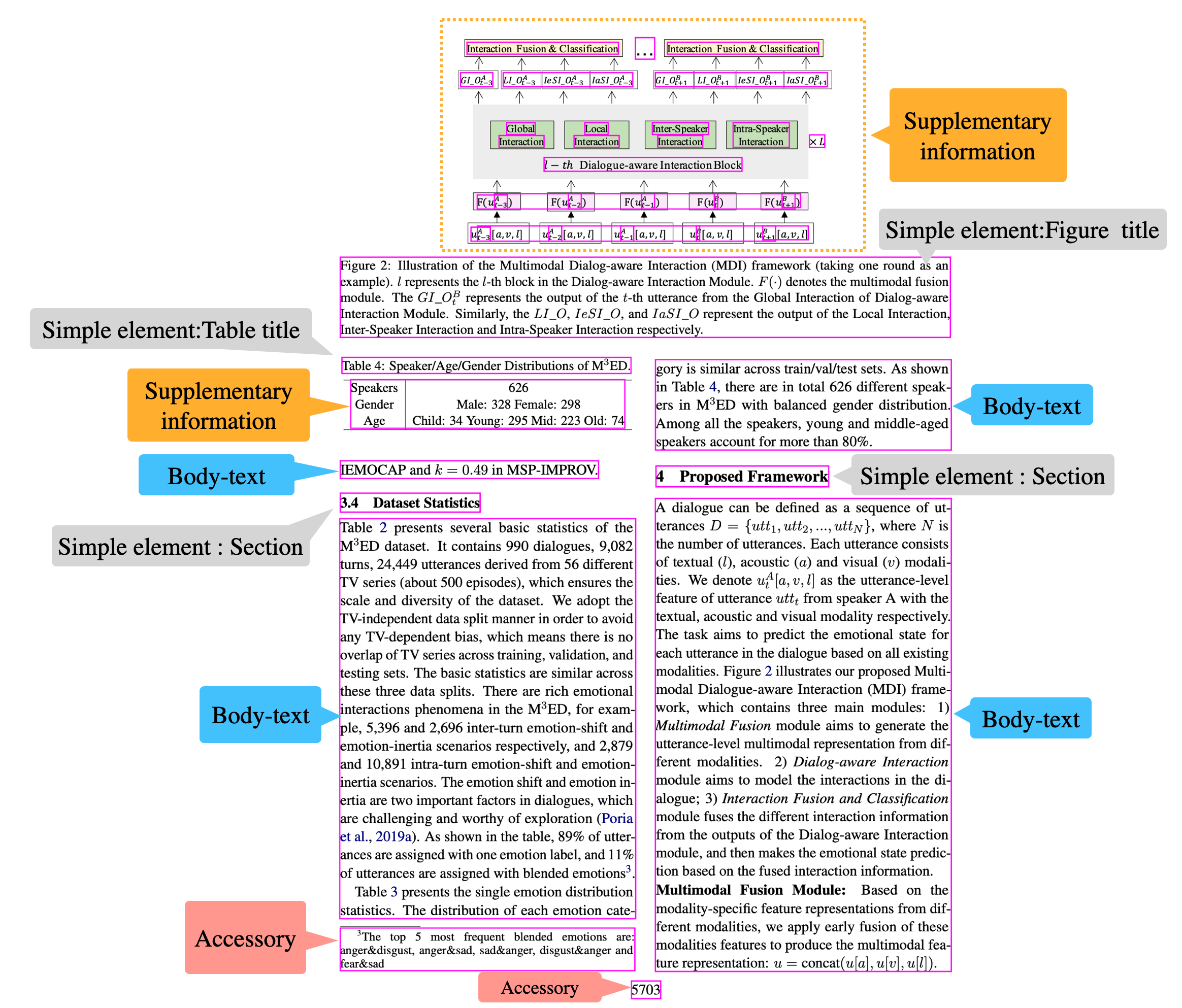}
\caption{Human annotation for text blocks\cite{fig4}.}
\label{f4}
\end{figure}

To enable the machine to recognize the types of information mentioned in Section \textbf{3.2.1} for each text block, we follow a process of encoding and vectorizing the text blocks. These text blocks, which contain multi-modal information, are manually annotated by humans. By combining the vectors of the constructed text blocks, we create a dataset for text box classification. This dataset has short annotation time, easy judgment, and high accuracy. An example of the labels can be seen in \textbf{Fig.\ref{f4}}.

\subsubsection{Classifier}
Support Vector Machine (SVM) is one of the classifiers that boast particularly high accuracy in machine learning. SVM has the advantage of high applicability even to unknown data\cite{SVM}. It seeks to maximize the margin when drawing a boundary line to divide the training data. Maximizing the margin improves the classification accuracy for unknown data, and the generalization performance becomes higher. Therefore, SVM has high adaptability and versatility for scientific documents with complex structure combinations. For instance, some special cases of text blocks in figures have the same font and size as the body-text. The SVM classifier’s margin and tolerance range setting are considered to improve training effectiveness to recognize these cases.

\subsection{Phase3 : Compartment Segmentation\& Object Recognition}
Using the text block classification results from Phase 2, we aim to cluster the classification results of text blocks into group levels for compartment segmentation and object recognition. Initially, we use single-modal text blocks such as section titles, figure titles, and table titles to establish the boundaries (refer to \textbf{Fig.\ref{f5}}). The content between these single-modal text blocks (for double-column layout documents, page break coordinates need to be set) may represent a compartment containing body-text or a compartment containing figures and tables. By combining the classification results of multi-modal text blocks, we statistically determine the category of the text blocks within this compartment and identify the type of object they represented. Finally, we establish the correlation between figure/table titles and their respective compartment based on their positions.

\subsubsection{Boundary setting based on simple text block}
\begin{figure}[htbp]
\centering
\includegraphics[width=14cm,height=10cm]{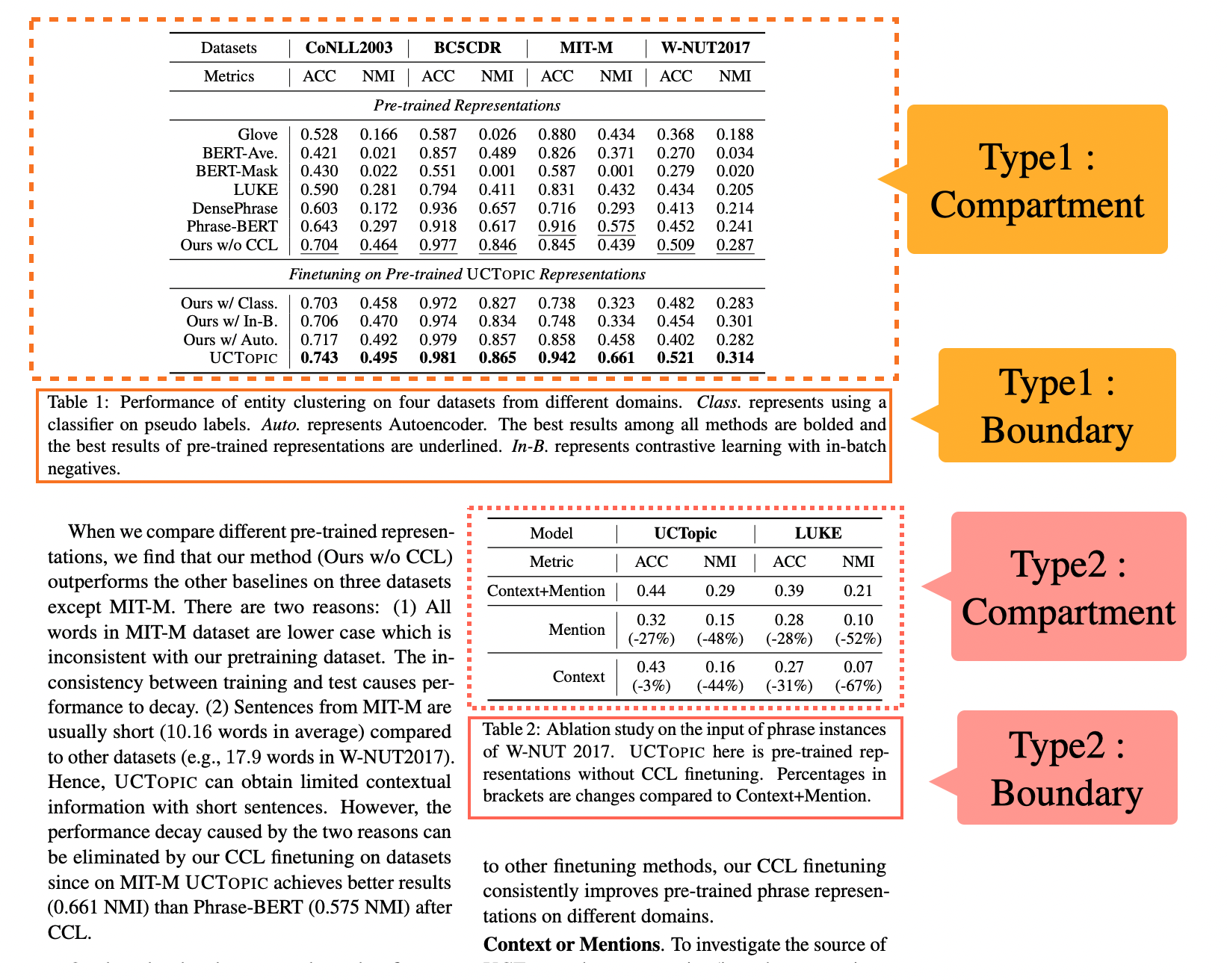}
\caption{Boundary setting \& compartment\cite{fig5}.}
\label{f5}
\end{figure}

We analogize the boundary lines in the document to doors or walls separating independent rooms. These boundary lines divide the entire document into several rough compartments. We determine the boundaries by considering the position and order of simple text blocks in the document, as illustrated in \textbf{Fig.\ref{f5}}.

\begin{itemize}
  \item \relax \textbf{Type1: Figure \& Table Title Crossing the Central Axis}
\end{itemize}
  If the text block area corresponding to the title of a figure or table crosses the central axis of the document, the region of the figure/table will also cross the central axis of the document. 

\begin{itemize}
  \item \relax \textbf{Type2: Figure \& Table Title Not Crossing the Central Axis}
\end{itemize}
  Conversely, if the text block area corresponding to the title of a figure or table does not cross the central axis of the document, then the region of the figure or table will follow the dual-column layout.

\subsubsection{Sophisticated Compartment Segmentation\& Object recognition}
In this section, we utilize the classification results of the text blocks and the position of the boundary to enhance the refinement of the rough compartment. We begin by assigning sequential numbers to the rough compartment and the boundary based on their arrangement in the document. Next, we iterate through the sequence of appearance of the figure and table titles that correspond to the boundary. Based on this strategy, we can infer the position and size of the compartments where the figures and tables are located, following the rules outlined below. This approach enables more precise compartment recognition.

\begin{figure}[htbp]
\centering
  \begin{minipage}[b]{0.45\linewidth}
    \centering
    \includegraphics[width=5.5cm,height=7.5cm]{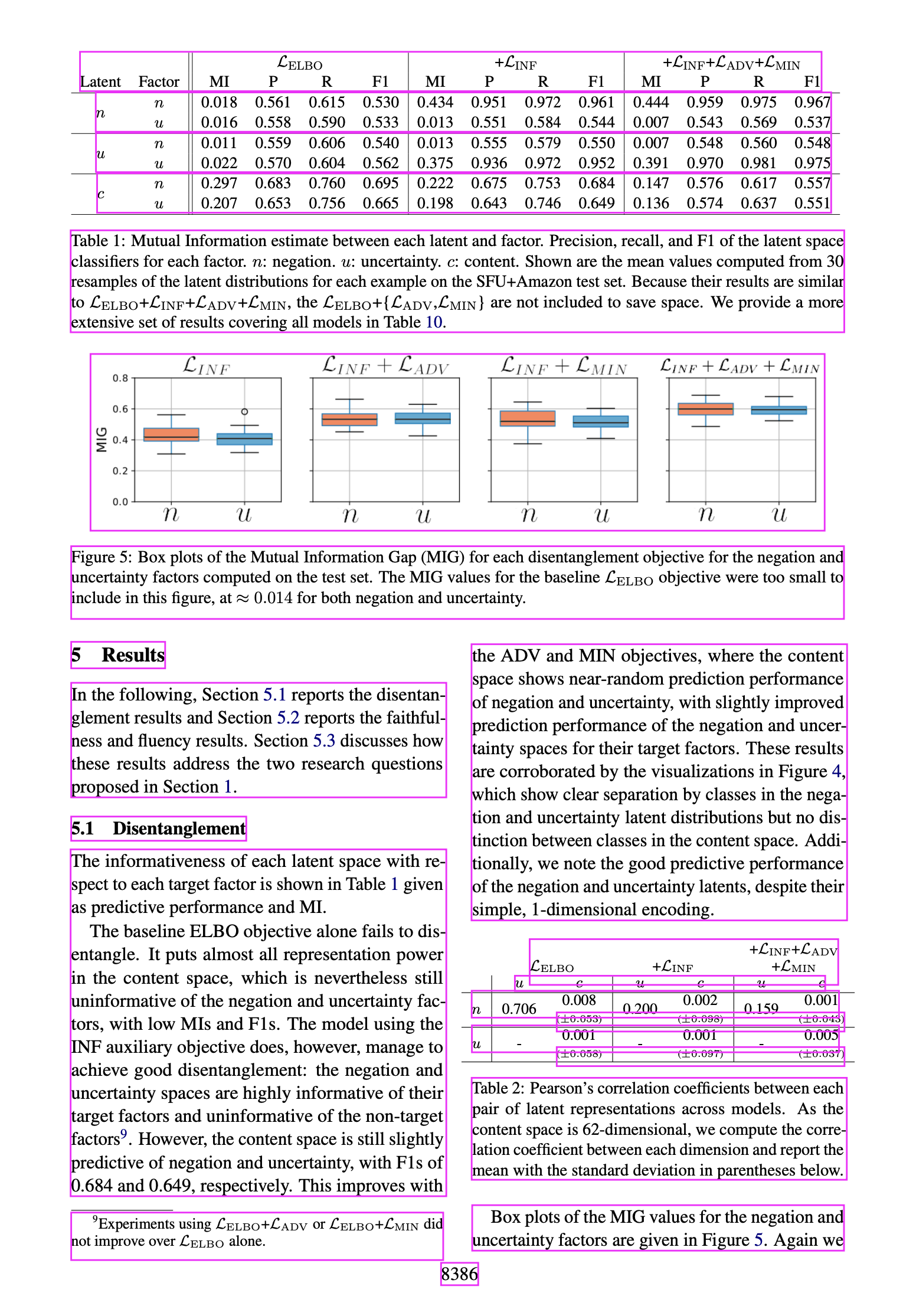}
    \subcaption{Unaligned layout and text block}
  \end{minipage}
  \begin{minipage}[b]{0.45\linewidth}
    \centering
    
    \includegraphics[width=5.5cm,height=7.5cm]{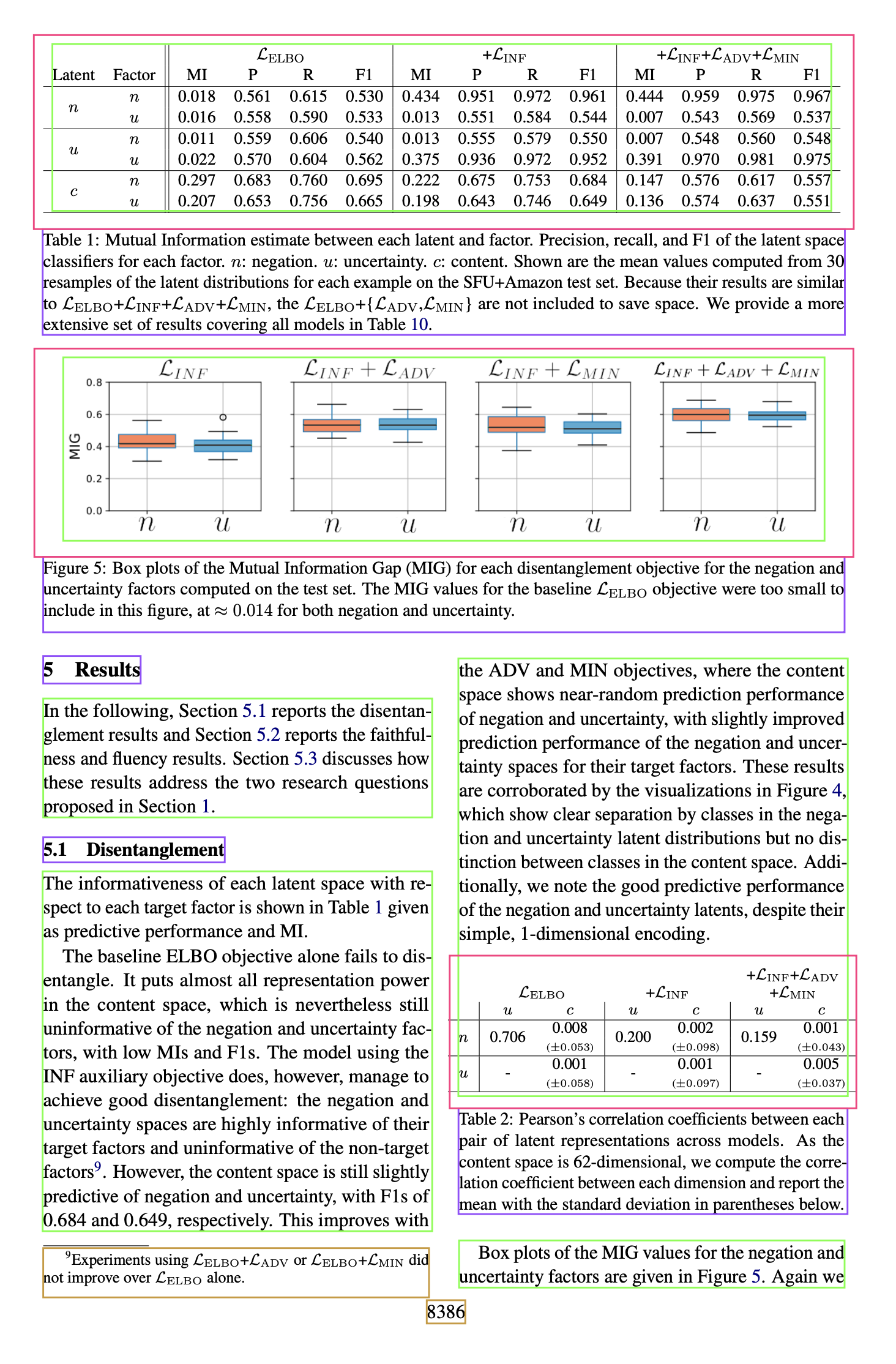}
    
    \subcaption{Figure,Table region detection}
  \end{minipage}
  \caption{Sample of Object Detection: Sample article\cite{detect}}
  \label{f6}
\end{figure}

  \begin{enumerate}
  \item \relax If the boundary where the figure/table title is located is at the top of the page (without any compartments appearing before it), the compartment where the figure or table is located must be directly below the boundary.
  \item \relax If the boundary where the figure/table title is located is at the bottom of the page (without any compartments appearing after it), the compartment where the figure or table is located must be directly above the boundary.
  \item \relax If the boundary where the figure/table title is located is in the middle of the page (with compartments before and after it), we need to use the text block classification results from the rough compartment, following Equation (7.1 - 7.3), to infer whether the corresponding figure/table region appears directly above or below. In Equation (7.1 - 7.3), \textbf{\textit{S}} means area.

\let\saveeqnno\theequation
\let\savefrac\frac
\def\dispfrac{\displaystyle\savefrac}
\begin{eqnarray}
\let\frac\dispfrac
\gdef\theequation{7.1}
\let\theHequation\theequation
\label{dfg-c58979722c05}
\begin{array}{@{}l}\begin{array}{l}Above\vert Below=Area\;of\;label\;'Supplementary'\;Occupied\\\end{array}\end{array}
\end{eqnarray}
\global\let\theequation\saveeqnno
\addtocounter{equation}{-1}\ignorespaces 
\vskip-1.5\baselineskip 
\let\saveeqnno\theequation
\let\savefrac\frac
\def\dispfrac{\displaystyle\savefrac}
\begin{eqnarray}
\let\frac\dispfrac
\gdef\theequation{7.2}
\let\theHequation\theequation
\label{dfg-eb445eb98a4b}
\begin{array}{@{}l}\begin{array}{l}Above\vert Below=\frac{\displaystyle\sum_{}^{}S_{block\_label\;of\;'Supplementary'}}{S_{Compartment}}\\\end{array}\end{array}
\end{eqnarray}
\global\let\theequation\saveeqnno
\addtocounter{equation}{-1}\ignorespaces 
\vskip-1.5\baselineskip 
\let\saveeqnno\theequation
\let\savefrac\frac
\def\dispfrac{\displaystyle\savefrac}
\begin{eqnarray}
\let\frac\dispfrac
\gdef\theequation{7.3}
\let\theHequation\theequation
\label{dfg-10ca03539c60}
\begin{array}{@{}l}Compartment\;=\left\{\begin{array}{l}Above>Below\;,\;Above\;is\;Figure/Table\;region\\Above<Below\;,\;Below\;is\;Figure/Table\;region\end{array}\right.\end{array}
\end{eqnarray}
\global\let\theequation\saveeqnno
\addtocounter{equation}{-1}\ignorespaces 
\item \relax Once the boundary-compartment correspondence is confirmed, other boundaries cannot occupy the same compartment.

\item \relax Since some compartments may contain additional information such as body-text or footnotes, we need to use the text block classification results for refining the rough compartment, following Equation (8.1 - 8.3) to further segment the figure \& table region.
\let\saveeqnno\theequation
\let\savefrac\frac
\def\dispfrac{\displaystyle\savefrac}
\begin{eqnarray}
\let\frac\dispfrac
\gdef\theequation{8.1}
\let\theHequation\theequation
\label{dfg-cdf819a07f90}
\begin{array}{@{}l}Bbox_{compartment}=(left_{pos},\;right_{pos},\;up_{pos,}\;bottom_{pos}\;)\end{array}
\end{eqnarray}
\global\let\theequation\saveeqnno
\addtocounter{equation}{-1}\ignorespaces 
\vskip-1.5\baselineskip 
\let\saveeqnno\theequation
\let\savefrac\frac
\def\dispfrac{\displaystyle\savefrac}
\begin{eqnarray}
\let\frac\dispfrac
\gdef\theequation{8.2}
\let\theHequation\theequation
\label{dfg-565a9ace6091}
\begin{array}{@{}l}Left_{pos}\vert Top_{pos}=Min\overset{}{\underset{}{{}_{Left\vert Top}\sum}}\;Text\;block\;of\;'supplementary'\end{array}
\end{eqnarray}
\global\let\theequation\saveeqnno
\addtocounter{equation}{-1}\ignorespaces 
\vskip-1.5\baselineskip 
\let\saveeqnno\theequation
\let\savefrac\frac
\def\dispfrac{\displaystyle\savefrac}
\begin{eqnarray}
\let\frac\dispfrac
\gdef\theequation{8.3}
\let\theHequation\theequation
\label{dfg-524aff780de1}
\begin{array}{@{}l}Right_{pos}\vert Bottom_{pos}=Max_{Right\vert Bottom}\sum Text\;block\;of\;'supplementary'\end{array}
\end{eqnarray}
\global\let\theequation\saveeqnno
\addtocounter{equation}{-1}\ignorespaces 

\item \relax Referring to the pdffigure2.0 method\cite{pdffigure}, if the width of the figure/table compartment is larger than the figure/table title, it is determined as the figure/table compartment. If the block's width where the figure/table title is located is larger, resize the figure/table compartment to match that width. The results of the Compartment recognition samples are shown in \textbf{Fig.\ref{f6}}.
\end{enumerate}

\section{Conclusion}
In this study, we developed an advanced framework called CTBR for text block classification and object recognition in scientific documents. This framework first defines the hierarchical structure of scientific document’s layout, including base domains, compartments, and text blocks. We developed a bottom-up level encoded template to refine these text blocks, which contain multi-modal data, and performed text block classification using machine learning and rule-based methods. Furthermore, we achieved more advanced compartment segmentation and object recognition using the classification results. The effectiveness of this framework was demonstrated through hierarchical scientific document layout analysis, using a small-scale training dataset and an SVM classifier for text block classification. We also developed a specialized algorithm for compartment segmentation to determine the region of figures and tables based on the classification results of text blocks, achieving an accuracy of over 90\%. Overall, the experimental results showed the effectiveness of this framework.\\

\textbf{Future tasks to improve this framework include the following:}

\begin{itemize}
\item \textbf{1. More refined compartment planning algorithm:}\\
This study divides the sections of figures, body-text, and footnotes in scientific documents. One of the future challenges is to add a finer level of division within the body-text, such as recognizing compartments of equations, itemized forms, algorithm areas, and lists.\\
\item \textbf{2. Compartment Internal Functional Differentiation}\\
In this work, we acknowledge the importance of the compartment in scientific documents. However, the function within these compartments is also crucial for a comprehensive understanding of the document. For example, the text blocks in figures can be considered components, each with a special meaning. In a statistical graph, the data on the horizontal and vertical axes reflect the range of the object, while the names of the axes indicate the measured indicators and so on. Similarly, the components of a model diagram in a scientific document provide a clear representation of the input-output and basic logic of the model. Exploring such detailed information further can optimize the dataset for scientific document understanding and greatly enhance the interpretability of figures in scientific documents.\\

\end{itemize}

\end{document}